# A Paradigm Change for Formal Syntax: Computational Algorithms in the Grammar of English

Anat Ninio

The Hebrew University of Jerusalem

## Abstract

Language sciences rely less and less on formal syntax as their base. The reason is probably its lack of psychological reality, knowingly avoided. Philosophers of science call for a paradigm shift in which explanations are by mechanisms, as in biology. We turned to programming languages as heuristic models for a process-based syntax of English. The combination of a functional word and a content word was chosen as the topic of modeling. Such combinations are very frequent, and their output is the important immediate constituents of sentences. We found their parallel in Object Oriented Programming where an all-methods element serves as an interface, and the content-full element serves as its implementation, defining computational objects. The fit of the model was tested by deriving three functional characteristics crucial for the algorithm and checking their presence in English grammar. We tested the reality of the interface-implementation mechanism on psycholinguistic and neurolinguistic evidence concerning processing, development and loss of syntax. The close fit and psychological reality of the mechanism suggests that a paradigm shift to an algorithmic theory of syntax is a possibility.

## Keywords

**1. Language Sciences are Abandoning Formal Syntax**

We are hearing more and more lately that the models of syntax offered by present-day theoretical linguistics are felt to be unusable by the client disciplines and applied fields dependent on it (de Bot, 2015). As typical exemplars of state-of-the-art formal theories of syntax we take Chomsky's Generative Grammar with its most recent version the Minimalist Program (Chomsky, 1995), and Dependency Grammar, in Hudson's version of Word Grammar (Hudson, 2010). As linguistics' applied disciplines, we refer to cognitive science, psycholinguistics, developmental psycholinguistics, applied linguistics, neurolinguistics, brain science, natural language processing, artificial intelligence, and so on. These disciplines are dependent on linguistics to supply the basic science for their own functioning.

The estrangement between the applied fields and the basic science which should serve them has been noticed, and from time to time it is the topic of an explicit discussion. Edelman and Christiansen (2003) summarizes the phenomenon in the journal "Trends in Cognitive Sciences":

[…] Minimalism is not even mentioned in recent reviews of, and opinions on, various aspects of language research in this journal ranging from sentence processing and production [5-7] and syntactic acquisition [8,9] to the brain mechanisms of syntactic comprehension [10-12]. We believe it would be in the best interests of linguistics and of cognitive science in general if the linguists were to help psychologists like ourselves to formulate and sharpen the really important foundational questions, and to address them experimentally. This, we think, would help cognitive scientists take Minimalist syntax more seriously. (p. 61).

We can add the situation in some other fields. IBM's Watson project made headlines with its computer program beating human competitors in the game of Jeopardy. This represented a considerable achievement in the automatic language understanding of natural speech, and especially a credit to the dependency-grammar based syntactic parser used, English Slot Grammar of McCord (McCord, Murdock, & Boguraev, 2012) on which it built a semantics-constructing component. However, in its recent crowning achievement, namely Project Debater, the language-understanding technology does not involve a parser. Its processing of the English sentences it is addressed with, and the ones it generates is given over to Artificial Intelligence or machine learning, employing purely statistics-based processes such as *data mining* that do not apply theoretical linguistic concepts and models (Slonim et al, 2021). In ten years, IBM's Watson has abandoned the use of the parser technology and with it, the potential ability to understand and generate meaningful sentences, with the loss of functionality this implies. Google has gone the same route.

Theory-based Natural Language Processing is no exception from this trend. A widely popular annotation scheme for syntactic dependencies, Universal Dependencies (Nivre et al., 2016), knowingly breaks with consensual linguistic opinion in treating all function words (such as auxiliary verbs) as dependents of the most closely related content word (i.e., lexical verbs), explicitly acknowledging that they are aware of disregarding expert opinion. As this decision affects the most frequent group of words in the relevant languages (e.g., close to 50% of all words in an English sentence), it has rippling effects on all the basic and applied disciplines using the annotation scheme. In particular, using this technique for studies in basic science such as exploring the concept of dependency distance (Liang, Fang, Lv, & Liu, 2017), or for classification of languages into head-initial or head-final (Gerdes, Kahane, & Chen, 2021), basically invalidates the results. In addition, employing this annotation scheme makes the relevant parsers less efficient than they would be with the linguistics-recommended annotation (e.g., Rehbein, Steen, & Do, 2017), thus directly harming the field of Natural Language Processing.

In the study of language acquisition (i.e., developmental psycholinguistics) the situation is a clear estrangement of the majority of the discipline from formal syntactic theories of the Chomskian and the Dependency Grammar types. In the recent Encyclopedia of Language Development (Brooks, Kempe, & Golson, 2014) there is no mention of studies employing the Minimalist Program, and the essay on theories of syntactic development in the paradigm of Dependency Grammar barely lists a few names of researchers who work in this tradition (Ninio, 2014). Instead of developmental theories that attempt to explain how children come to possess the knowledge described by linguistics, the field drifts to either of two directions. Some model syntactic development as statistical learning of distributional regularities (Bannard, Lieven, & Tomasello, 2009; Saffran, 2003, and see also Christiansen & Chater, 2008 for a review), continuing a line of research proposing for young children the acquisition of sequential formulae such as pivot-open combinations (Braine, 1963), or formulae consisting of a fixed frame with variable slots (Pine & Lieven, 1993), their components defined purely by their distributional features. Unfortunately, these alternatives to formal models of syntax suffer from serious weaknesses of their own (see Ninio, 2011, pp. 10-13). It is fundamentally impossible to exchange hierarchical syntactic structure with sequential analyses.

The second trend is to adopt some version of the heavily semantically-oriented Construction Grammar (Goldberg, 1995, Croft, 2001) that does no hold the basic principle of formal grammars regarding the autonomy of syntax from semantics. Instead, the units of sentence structure are said to be constructions which are pairings of linguistic patterns with prototypical meanings. However, this circumvention of formal syntax does not fare well either. It is well known that the core syntactic relations -- the subject-verb, verb-object and verb-indirect object relations -- which young children learn at the start of acquisition -- possess an unusually wide semantic range, and said to be semantically neutral (Andrews, 1985; Givón, 1997). Thus, in English parental speech the most frequent verbs in the basic transitive patterns of verb-direct object and subject-verb-direct object do not have prototypical transitive semantics such as "X acts on Y" (Goldberg, 1999; Sethuraman & Goodman, 2004), and young children's earliest transitive sentences ("*Want bottle*") have stative semantics, not transitive ones (Bowerman, 1978). These findings call for a return to formal syntax in models of language development. That, however, appears to be more difficult than possible.

## 2. Formal Syntax Lacks Psychological Reality

It is should not surprise us that cognitive science, developmental psycholinguistics, natural language processing and so on do not use present-day formal syntactic models as their theoretical umbrella, because the latter are not meant to be so used in the first place.

Let us take Chomsky's Generative Grammar as our example of a very prominent theory of syntax and see how it was explicitly and openly not intended to be usable as a model of human processing of speech.

In his very earliest publications, Chomsky (1957) describes the task of a grammar as the generation of the grammatical sentences of a language:

The fundamental aim in the linguistic analysis of a language L is to separate the *grammatical* sequences which are the sentences of L from the *ungrammatical* sequences which are not sentences of L and to study the structure of the grammatical sequences. The grammar of L will thus be a device that generates all of the grammatical sequences of L and none of the ungrammatical ones. (p. 13)

It goes without saying that the goal of these grammars was to be elegant, that is, demonstrate "the infinite use of finite means" (Chomsky, 1965, p. v, citing von Humboldt, 1836). This was indeed achieved by the generative grammars of these earlier versions, where the generation of an infinite number of sentences by the grammar was carried out by a few so-called *rewrite rules* by which a symbol of a larger unit on the left is rulefully rewritten by its grammatical constituents on the right. Such rules start from the top, with the symbol for *Sentence*. The rewriting continues until every constituent unit reaches its lower limit in the form of word-level, non-phrasal or terminal categories such as Noun, Verb, Adjective, Determiner and so forth. For each sentence it is possible to derive a *phrase marker* based on the actual steps of its derivation. The phrase marker represents the structure of the sentence in abstract terms; a further step of *lexical insertion* substitutes concrete words for the terminal symbols of the phrase marker to generate an actual sentence. Let us see the earliest description of such rules in Chomsky (1957):

As a simple example of the new form for grammars associated with constituent analysis, consider the following:

(13) (i)         *Sentence* $\rightarrow$ *NP + VP*

    (ii)         *NP* $\rightarrow$ *T + N*

    (iii)         *VP* $\rightarrow$ *Verb + NP*

    (iv)         *T* $\rightarrow$ *the*

    (v)         *N* $\rightarrow$ *man, ball*, etc

    (vi)         *Verb* $\rightarrow$ *hit, took*, etc.

Suppose we interpret each rule $X \rightarrow Y$ of (13) as the instruction "rewrite $X$ as $Y$". We shall call (14) a *derivation* of the sentence "the man hit the ball" where the numbers at the right of each line of derivation refer to the rule of the "grammar" (13) used in constructing that line from the preceding line. (pp. 26-27)

This sounded to psychologists of the time like the description of a cognitive process, for example, it was thought that the more derivational steps it takes to generate a sentence, the more difficult the processing and learning of such sentences would be. See Watt (1970) for an early and comprehensive discussion of the hypothesis of derivational complexity and its negative results. Chomsky (1965) felt the need to explain that he only meant by "generate" the attribution of a linguistic description by the grammar, not psychological processes performed by speakers:

To avoid what has been a continuing misunderstanding, it is perhaps worth while to reiterate that a generative grammar is not a model for a speaker or a hearer. […] When we speak of a grammar as generating a sentence with a certain structural description, we mean simply that the grammar assigns this structural description to the sentence. (p. 9)

Less the point be not fully understood, later in the same monograph he explains in detail why it is absurd to regard his rules of derivation as referring to the actual cognitive processes carried out during the construction of a sentence by a speaker:

Such a description of the form of the syntactic component may seem strange if one considers the generative rules as a model for the actual construction of a sentence by a speaker. Thus it seems absurd to suppose that the speaker first forms a generalized Phrase-marker by base rules and then tests it for well-formedness by applying transformational rules to see if it gives, finally, a well-formed sentence. But this absurdity is simply a corollary to the deeper absurdity of regarding the system of generative rules as a point-by-point model for the actual construction of a sentence by a speaker. Consider the simpler case of a phrase structure grammar with no transformations (for example, the grammar of a programming language, or elementary arithmetic, or some small part of English that might be described in these terms).
It would clearly be absurd to suppose that the "speaker" of such a language, in formulating an
"utterance," first selects the major categories, then the categories into which these are
analyzed, and so forth, finally, at the end of the process, selecting the words or symbols that
he is going to use (deciding what he is going to talk about)." (pp. 139-140).

He returns to the point in his next book, categorically denying that the derivational acts of his
grammar are intended to describe something that actually happens in people's minds:

> [...] although we may describe the grammar G as a system of processes and rules that apply
> in a certain order to relate sound and meaning, we are not entitled to take this as a description
> of the successive acts of a performance model *PM* — in fact, it would be quite absurd to do
> so. (Chomsky, 1968, p. 117)

In the latest version, the Minimalist Program (Chomsky, 1995, p. 226), the derivation
proceeds from the bottom towards the top, namely, the words of the sentence merge with other
words or with symbols for already-combined, larger constituents in a binary combinational
system. Merging consists of the creation of a set consisting of the two merging components. As a
second, separate step of derivation, this set is then labeled with the form-class label of the head-
constituent. We do not need Chomsky to repeat his warning in order not to take these processes
as the intended descriptions of actions actually performed by speakers when they generate
sentences. In a paraphrase on his texts above, we might say that it would be absurd to think he
means to claim that when deriving sentences in reality, people first put two words into a set,
without combining them or ordering them in any way, and subsequently only label the
combination as some sort of phrase, the description still not mentioning a process that would
actually regulate the combination of the information of the two words as semantics requires.

Chomsky of course has legitimate theoretical or mathematical reasons for using such a description of the basic combinatory process of syntax, for which we cannot criticize him. However, there are linguists and psycholinguists who find this brand of theorizing unacceptable, for instance see Pullum's (2009) severe critique. For our purposes, the voices coming from the client-disciplines of theoretical linguistics are more relevant. Despite Chomsky being clear about not intending to claim psychological reality for his derivational processes, some psycholinguists complain that present-day linguistic theories cannot be taken seriously when there is no evidence that the operations they mention, e.g., merge and move, possess reality of any kind. Thus, Edelman and Christiansen (2003) argue for "the need to demonstrate the psychological (behavioral), and, eventually, the neurobiological, reality of theoretical constructs. […] Unfortunately, to our knowledge, no experimental evidence has been offered to date that suggests that merge and move are real (in the same sense that the spatial-frequency channels in human vision are)" (p. 60).

Similarly, Ferreira (2005) says:

In the 1980s, Charles Clifton referred to a "psycholinguistic renaissance" in cognitive science. During that time, there was almost unanimous agreement that any self-respecting psycholinguist would make sure to keep abreast of major developments in generative grammar, because a competence model was essential, and the linguistic theory was the proper description of that competence. But today, many psycholinguists are disenchanted with generative grammar. One reason is that the Minimalist Program is difficult to adapt to processing models. (p. 365)

The choice to construct grammars as a system of rules and derivations rather than of processes is not unique to Chomsky's Generative Grammar but, rather, a general feature of present-day formal theories of language. For instance, Hudson (2014) in his *Encyclopedia of Word Grammar and English grammar* describes his grammar as a collection of rules, and, similarly to Chomsky, warns us not to confuse rules with psychologically real processes:

A rule is just a fact; it may be very general or very specific, and there is no important difference between `rules' and `principles'. WG rules are not procedures - operations to be performed - but static descriptions of patterns that are acceptable (or unacceptable, if 'not' is allowed) - permissible combinations of words, or of word and meaning, and so on.

An apparent exception is Montague-type grammars that do aim for an operational description, but their atoms are concepts and the structure they describe is semantic, not syntactic (Bar-Hillel, 1953; Cann & Kempson, 2017).

Generative grammars are indeed impressive on their own terms. As promised, they show how an infinite number of grammatical sentences is generated with the help of a finite number of rules of permissible patterns. The problems start when we realize that what such models of syntax aim at is to excel in an intellectual game of creating elegant derivations that generate all and only grammatical sentences, without being bothered by the absence of the psychological reality of the describing concepts, rules and virtual actions. Their intended audience is other theoretical linguists who can appreciate the achievement for what it is, and not, unfortunately, the applied disciplines such as cognitive science and neurolinguistics. These expect to be provided by clear guidelines to a psychologically real syntax, that would tell them what actual syntactic computations consist of. For developmental psycholinguistics, a psychologically real syntax would tell what children have to learn to actually be able to generate and interpret sentences. This calls for a paradigm shift.

**3. Philosophy of Science Calling for a Paradigm Shift.**

The absence of a model of linguistic competence that refers to actual mental operations is a conspicuous exception in cognitive psychology. Other cognitive functions such as vision are already described as systems of operations, forming mechanisms taking in input stimuli and resulting in some mental representation of reality (Marr, 1982). Language is an exception, although most certainly it is a cognitive function. In this case, the syntactic structure of sentences is characterized by static descriptions of acceptable or permissible patterns of words. As we saw above, the dynamic operations referred to by present-day theories (such as movements) are the unreal variety, and do not attempt to describe actual processes.

Our plan for a paradigm shift for syntax that would refer to psychologically real cognitive processes is by no means a new idea. In actuality it follows decades-long recommendations coming from philosophy of science for a paradigm change for modeling cognitive phenomena. Authors such as Bechtel and Richardson (1993), Glennan (2002) and Machamer, Darden, and Craver (2000) suggest for cognitive sciences a shift away from a deductive-nomological model of scientific explanation which uses linguistic and mathematical representations for laws, initial conditions and derivations from them to yield the described phenomena (Bechtel, 2009, p. 553). These authors believe that the correct method to explain natural phenomena in general and cognitive phenomena in particular is by describing mechanisms that could produce the said phenomena. Newell and Simon (1972) represent an even earlier generation who proposed that psychological research should discover processes by which cognitive tasks are performed.

What in natural science and in particular in biology is called a mechanism, is in fact equivalent to what in mathematics, logic and computer science is entitled an algorithm. According to a succinct definition from the philosophical literature, "A mechanism for a phenomenon consists of entities (or parts) whose activities and interactions are organized in such a way that they produce the phenomenon." (Glennan 2017, p.17). An algorithm is defined very similarly, as an explicit and precise sequence of instructions for performing computational tasks (Berlinski, 2000; Schneider & Gersting, 1995). Importantly, an algorithm, like a mechanism, is a deterministic procedure; as Vinacke (2000) says, "A problem-solving algorithm is a procedure that is guaranteed to produce a solution if it is followed strictly." We shall be using both terms interchangeably.

Craver and Krickel (2019) emphasize that the search for mechanisms is a widely adopted method of science:

> Many areas of science are animated by the search for mechanisms: experiments are designed to find them, explanations are built to reveal them, models are constructed to describe them, funding is disbursed to prioritize their discovery, and translational research is premised on their value for manipulation and control. Over the last twenty or thirty years, a number of philosophers of science, starting from the biological and neural sciences, have directed attention at the concept of mechanism. They have emphasized that mechanisms play central roles in discovery, explanation, experimentation, modeling, and reduction. (p. 395)

## 4. Modeling on Programming Languages and Choice of Syntactic Pattern to Model

### *4.1. Turning to programming languages for a heuristic model of algorithmic syntax*

Judging from the field's history, the construction of syntactic theories necessarily involves the adoption of a formal language as the theory's framework, such as a system of logic or mathematics. The characteristics of the specific system chosen has a decisive effect on the resulting grammars. Bar-Hillel (1953) chose as his source the formal logic of Kasimir Ajdukiewicz, a Polish logician, resulting in his Categorial Grammar possessing a strong leaning to semantics and also suffering from an unsolved problem of natural languages' rampant category ambiguity. Similarly, Montague (1973) built his grammar on the basis of mathematical logic, adopting Alonzo Church's Lambda Calculus as his framework, which resulted in a semantic grammar which is ambiguously also a syntactic system. Other Montague-type grammars are based on similar kinds of logical formalization, for instance Steedman's (1996) Combinatory Categorial Grammar and Kempson, Meyer-Viol and Gabbay's (2001) Dynamic Syntax took Haskell Curry's Combinatory Logic as a frame, resulting in semantic systems that deny a separate level of syntactic structure.

In a departure from the semantic leanings of the former group, Chomsky (1959, 137n) took as his frame mathematical computability theory, adopting Emil Post's top-down random generator and his rewrite rules as the prototype of syntactic derivation, resulting in a system unlikely to have psychological reality. As Pullum (2009) pointed out: "…surely it is not a sensible hypothesis about human linguistic competence to posit that in the brain of every human being there is an internalized random generator generating every physically possible sequence of sounds, from a ship's foghorn to Mahler's ninth symphony (p. 14).

Our chosen way to a psychologically real model of syntax built of algorithms is to follow an established but lately somehow neglected path in theorizing: using the theory of computation as a source of ideas. Computer programming languages are obvious heuristic models for natural languages, with a better chance than other systems to result in a realistic outcome. Many researchers, starting with Montague's (1973), thought that natural languages such as English are fundamentally very similar to formal or artificial languages, and that it is possible to treat them with similar methods. Over the years there have been various suggestions that natural languages be seen as a kind of programming language forming a collection of instructions (e.g., Eijck &

Unger, 2010; Pylyshyn, 1972; Simon, 1979), but these ideas never matured to an actual theory of syntax modeled after the design of computer programs. Many of these proposals, including the recent Eijck and Unger (2010) and Kempson, Cann, Gregoromichelaki, and Chatzikyriakidis (2017), use a process-oriented method in order to directly model semantic rather than syntactic interpretation of the sentence, with Cann et al. claiming that syntax is an unnecessary level of theorizing sentence structure. In this work we shall continue to make the consensual assumption that creating sentence structure is a separate regulatory task belonging to syntax.

Recently, Wigderson (2019) invited researchers to integrate computational theory into natural and social sciences, believing this move will be at the heart of a new scientific revolution (p. 232). In the following, we shall attempt to implement Wigderson's vision in the development of an algorithmic model of syntax, using as our source of inspiration the devices of computer programming.

Whereas natural languages have evolved to reach their present shape and anything concerning their development and characterization is open for speculation, programming languages are manmade formal languages, meaning they are totally explicit and transparent. When using them as heuristic models, we can benefit from their extensive documentation in which their architecture and design is clearly described and analyzed.

**4.2. The main principles of our method of modeling**

A computer program, or software program, is a set of instructions that directs a computer to perform a task. The kind of program we are interested in is the one human programmers write, using a *high-level* or human-readable programming language (or *assembly language*) such as Python or Java. Direct control of the central processing unit of a computer requires an equivalent program written in machine code (mostly a sequence of zeroes and ones), to which the high-level program is translated by a utility such as a compiler. High-level programs have to be completely explicit in anything that they tell the computer to do, and they have to follow strict formatting rules. This allows interested audiences to follow the programmers' strategies and the algorithms they employ. In addition, programming languages are extensively described in textbooks as well as discussed in dedicated web sites such as "GeeksforGeeks" or "StackOverflow", making it possible to find out the reasons why the various devices of the languages were originally developed and how they continue to evolve. Overall these documentations provide transparent information about the makeup of the various programming languages, making it relatively easy to use them as heuristic models for natural languages in the way we are using them.

For the purposes of our modeling, we shall consider each English sentence the equivalent of a complete program, whose task is to compute the sentence's global meaning. Each word, with its lexically stored valency, will be taken as the parallel of a single programming statement or declaration in that program. Our unknown and to-be-discovered information are the processes by which two words combine to form a higher-level structural unit, which we shall call for simplicity's sake by their usual labels a phrase, a clause or a sentence. Our algorithms, then, are the step-by-step instructions by which structural units are computed by psychologically real cognitive operations.

In the following, we shall use some word-combinations of Modern English as the data on which to develop the outline of an algorithmic syntax. As our foundation for model-building we shall take from the vast linguistic literature treating English everything which is descriptive, theory-neutral, and consensual. That is, we shall take as given the terms, rules and regulations of English documented in traditional or descriptive grammars such as Bloomfield (1933), Bolinger (1975), Hockett (1958), Hornby (1954), Huddleston and Pullum (2002), Lyons (1968), and in particular Quirk, Greenbaum, Leech, and Svartvik (1972). In addition, we shall accept the descriptive components of generative grammars as well, for example of Haegeman (1994) and Hudson (1990). That means we shall be using the traditional classification of the vocabulary to form-classes, so that we shall be speaking about nouns, verbs, determiners and relative pronouns and so forth, mostly for the sake of easy communication. We shall define syntax in the traditional way, as the combination of words, not of bound morphemes or unexpressed conceptual categories. As it is routine in present-day theories, we shall take direct syntactic relations as binary, namely involving, at the bottom, the combination of two words. We shall accept the customary identification of some word-combinations as syntactic and others as not, so that in the phrase *a young boy* we shall treat the words *a* and *boy*, and also *young* and *boy* as in a direct syntactic relation, whereas *a* and *young* are not. In particular, we shall use the distributional data presented in grammars in the forms of rules of form and word order of word-combinations. Lastly, as we said above, we shall use the conventional terminology for the outcomes of word-combinations, namely call some combination a verb phrase, a noun phrase or determiner phrase, a subordinate clause, a relative clause, or similar. What we shall not accept are the theoretical derivational routes by which the combination of words is claimed to arrive at the relevant result.

Our modeling strategy, then, consists of the following. Given a pair of words which, according to present grammars, are in an acknowledged syntactic relation, and given the result of their combination, again according to present grammars, we shall search in the programming languages theory for two elements with analogous features combining to a similar outcome. Our goal is to learn in detail the computational process or algorithm that generates the said outcome from the two input elements in the relevant programming language. Our test for the success of our modeling is how well the linguistic features of the English word-combination fit the functional characteristics of the discovered computational algorithm. We will also test the psychological reality of the proposed cognitive process by examining psycholinguistic, developmental and neurolinguistic evidence on its effects in adult and child speakers of English. Our hope is that the identification of algorithms responsible for the syntactic structuring of sentences would satisfy the need of the related sciences and applied fields for a usable linguistic model of syntax.

### 4.3. The type of word-combinations chosen to model

As our exemplar of English syntax to reanalyze in terms of algorithms, we have decided to focus on the grammatical combinations of a function word (FW) and a content word (CW), producing phrases and clauses.

Content words (also lexical items) are nouns, verbs, adjectives and adverbs. Function words (also grammatical items, form words, structural words) include auxiliary verbs (e.g., *be, have, do*, and also infinitive-*to*), copulas (*be, become*, *turn*), modal verbs (*can, could, may, must*), determiners (*the, a, an, this, that, these*, *some*), subordinating conjunctions (*that, if, because*), prepositions (*in, on, to, for, from*), relative pronouns (*that, who, which*), and question words (*who, what, which, where, why*). Often adverbs of degree (e.g., *very, quite, more*) are also considered FWs. The FW vocabulary also includes coordinating conjunctions (*and, but, for, nor, or, so,* and *yet*), but their syntactic pattern is dissimilar to the other FWs', and we shall not discuss them further in the present paper.

Lyons (1968) summarizes the distinction between CWs and FWs: "Lexical items are traditionally said to have both 'lexical' ('material') and 'grammatical' ('formal') meaning. Grammatical items are generally described as having only 'grammatical' meaning" (p. 438). Thus, according to Lyons' definition, the FW-CW combination is the combination of an all-grammar word with a word possessing grammar-plus-semantic content. Table 1 presents the major kinds of such combinations by the form class of the combining elements and the outcome of their combination, with some examples.

Table 1

*Word combinations of FWs and CWs, and their outcome*

---

| Form class of FW | Form class of CW | Outcome | Examples |
|---|---|---|---|
| Auxiliary verb | Nonfinite lexical verb | Verb phrase | *will arrive; can open* |
| Copular verb | Nonfinite predicate | Verb phrase | *is happy; turned twenty* |
| Determiner | Common noun | Determiner phrase | *a boy; some advice* |
| Adverb of degree | Adjective | Adjective phrase | *very nice; quite tall* |
| Adverb of degree | Adverb | Adverb phrase | *very fast; more carefully* |
| Subordinating conjunction | Finite verb | Subordinate clause | *that they are not coming; if everything goes right* |
| Relative pronoun | Finite verb | Relative clause | *whom I like the best; where Mary sits* |
| Question word | Finite verb | Interrogative clause | *Who is she? Why are you so late?* |

---

Such combinations are the ideal exemplars to serve as the basis of a new model of English syntax, for two reasons. First, they are probably the most frequent type of word-combinations in English, together accounting for close to 50% of all syntactic combinations in texts (Ferrer-i-Cancho & Solé, 2001). Second, these deceptively simple word-combinations are quite important in English syntax, given that they are responsible for producing the major constituents of sentences. If we return to Chomsky's (1957) earliest definition of rewrite rules cited above in Section 2, we find that the very first derivational rule is *Sentence* → *NP + VP,* meaning that sentences are to be constructed of NPs (noun phrases) plus VPs (verb phrases). Further rules tell us what are NPs constructed of: *NP* → *T + N* , when *T* is the article *the* and *N* is a noun such as *man* or *ball*. Although this miniature grammar was used merely as an example of what Chomsky called "the new form for grammars associated with constituent analysis", these rewrite rules express the structuralist insight that the immediate constituents of sentences are noun phrases and verb phrases (Bloomfield, 1933), and that the construction of these phrases very often involves function words such as the article *the* combining with a content word such as the noun *man*. This generalization is true for all syntactic FW-CW combinations considered here.

Despite their importance, the derivational account of such combinations does not shed light on the actual processes by which such combinations "project" to phrases and clauses. The derivational account describes a bipartite or double "projection" in which a "functional shell surrounds the lexical projection" (Grimshaw 2000, p. 116). Other descriptions talk about multiple zones and embeddings and similar multilevel structures, but not actual syntactic processes beyond static depictions. As we reminded ourselves, such descriptions do not intend to refer to psychologically real cognitive mechanisms in which the two input elements interact to produce phrases or clauses.

In some morphological theories associated with generative grammar, FW-CW combinations are conceived as word-formation processes in which words, seen as originally class-less roots or *listemes*, are only specified for a semantic and syntactic category such as *verb* or *noun* by the slot in the syntactic context provided by the FW (Borer, 2005; Marantz, 1997). This possibility does not conform with the empirical evidence by which words "partially converted" by syntactic content nevertheless keep their original inflectional behaviour, and moreover do not acquire the morphological features of their acquired form-class (Quirk et al., 1972, p. 1010). As Quirk et al. summarize it, it is doubtful whether the syntactically induced categorization should be treated as a word-formation process at all. Contemporary publications present further counter-evidence to these ideas, see for example several chapters in Alexiadou, Borer, and Schäfer (2014). In other texts discussing the FW-CW combination, we find descriptions of various non-syntactic processes such as marking, cuing, signalling or predicting, that are said to describe the interaction between the two types of words or to account for the generation of the higher syntactic units (e.g., Ferreira & Qiu 2021, Hawkins, 2004: Vogel, 2000). Pointing out the occurrence of a statistical, correlational relationship between input items, or between them and the output provides valuable information about the relevant pattern, but does not constitute a theoretical model of it. The development of such a model in algorithmic terms is the goal of the present work.

In the following, we shall report on a programming algorithm with inputs and outputs analogous to the FW-CW combination, the interface-implementation mechanism (section 5); next, we test the English pattern's fit to the functional features of this algorithm (section 6); test the algorithm's psychological reality in syntactic processing, development and loss (section 7); and discuss the significance of the results (section 8).

## 5. A Programming Algorithm Producing Typed Objects

In Section 4 we decided to explore the syntactic combination of a function word (FW) and a content word (CW) in English that produces a phrase or clause. In more abstract terms, this pattern consists of two lower-level input elements whose output is a higher-order typed unit.

We now search for the analogue of the same input and output pattern in programming languages in order to gain an algorithmic or mechanistic explanation for the pattern. The gain will be understanding how the interaction of these inputs produces the output. We have criticized present-day generative grammars for describing the derivation of sentences in terms of unreal processes such as *projection*, *enveloping*, or *movements*. In programming, input-output relations analogous to the patterns observed in natural sentences will be the outcomes of meaningful computational processes. Our goal is to learn what such computational algorithms consist of in programming languages, and to test if their analogue fits natural languages.

In the present case, we have focused on FW-CW combinations that generate phrases or clauses such as a noun-phrase or a subordinate clause. The latter are immediate constituents of English sentences, and understanding how they are arrived at in computational terms would be a large step towards understanding what psychologically real derivation of sentences consists of.

We easily found in computer programming a two-step mechanism for the generation of computational objects; the parallel to the FW-CW pattern is the *interface-implementation* algorithm used in object-oriented programming (OOP) languages. In all programming languages, the data on which a program is to operate are specified by explicit definitions or declarations of each variable. Such definitions consist of a name and a data type for the relevant variable. The data elements or objects so defined are then available for the computational actions the program specifies to be performed on them. The algorithm we found is a method of declaration of some object or variable of a program, central in OOP. Unusually, it consists of a pair of elements first abstracting the gist of the object, then realizing the abstraction.

A few words on OOP. It conceptualizes a computer program as a complex system which is made up of *interacting objects* (Kay, 1993; Goldberg, 1983). This in contrast to function-oriented programming where a program is treated as a sequence of abstract function evaluations. In programming languages with an object-oriented design, such as Java, C++ or Python, objects are similar to real-life things and beings, in fact this kind of program originated in a self-conscious tendency to anthropomorphize the computational world (Cockburn, 1992). Objects in OOP are software entities that contain *data* (their characteristic properties or attributes and momentary state), and also procedures that access and modify the data, namely operations or functions that it performs or that can change its states, called its *methods* or *behaviours*.

The relevant object-generating algorithm of OOP consists of defining two parts for an object, a public façade called *interface*, consisting of a few behaviours characteristic of such objects in general (its *public member functions*), and an internal space called *implementation* (or *body*) in which information on the momentarily relevant specific instance of that class, its private properties, states and operations are specified (Martin, 1993). In some programming languages the interface consist only of so called *signatures*, namely descriptions of behaviours, in others such as Java, they can be actual executable actions.

OOP explicitly separates the interface from the body of the object, following a principle called *abstraction* (Booch et al., 2007) and sometimes simply *interface*. The separation is physical or corporeal: each of these component is a separate entity. We may say that objects in OOP are complex; they consist of two objects with complementary roles. The interface element forms the first statement introducing the object, hence literally consisting of the object's border (Gamma, Helm, Johnson, & Vlissides 1994). All communication with the object is through the interface, hence the internal complexity is irrelevant for operations towards the object by external elements.

As we mentioned above, by OOP a program consists of objects interacting with each other, asking for certain computations or data. More precisely, the component of objects that is accessible for other objects is the interface element. Communication between objects means, therefore, communication with their interface, which is their sole visible part.

The interaction between objects is conceptualized in OOP as the relation between *clients* that have needs and *servers* that satisfy those needs. More formally, a client is an object that uses the resources of another object (Booch, Rumbaugh, & Jacobson, 2005). The streamlining of this relationship is a major goal of OOP (Booch et al., 2007). Objects send messages to other objects – when, essentially, a message is a procedure call from one object to another. They are sent to objects to trigger their behaviour or provide their data, which is the kind of service the client-object requires.

The interface is the part of the object shared with external objects, which they communicate with. It offers them what they may need and accepts from them a request to supply it. In other words, if we know what is contained in an interface, we can identify what is it that the client object is looking for.

Gamma et al. (1994) say that an object's interface characterizes the complete set of requests that can be sent to the object. We can turn this around and say that interfaces characterize clients' requests – this is what the client asks for. That is, separating the interface from the implementation is based on the principle that the part shared with the client of the object contains exactly what client needs to know, no more and no less.

Given that the information in the interface is one that identifies the type of the object but not anything that gives away its properties and other unique details, we can assume that what the client requires is to know that an object of a given type will be coming, not anything more specific about it. That is, interfaces are the realizations of client needs in the form of essential behaviours for the relevant type of objects.

The question to ask now is how this mechanism works to generate the desired objects in programming. The answer is that the interface-implementation algorithm uses a programming gimmick called *duck typing*, used for instance in Python. Duck typing uses the so-called *duck test:* "If it looks like a duck, swims like a duck, and quacks like a duck, then it must be a duck". In formal terms, this is a kind of *abductive reasoning*. In terms of programming, duck typing identifies an unknown entity by its habitual characteristics and behaviour. In the interface-implementation algorithm, a new object is defined by the behaviours it must have according to the needs of the program. We might call it a duck-defining algorithm.

The truly unusual aspect of definition by characteristic behaviours (i.e., *duck typing*) is that it makes the categories of origin of the defined object irrelevant. In such a system, there are no class labels and no inheritance from classes of origin. The object can come from any category and it will be accepted as long as it conforms to the definition by having the correct behaviours.

If the interface part of the algorithm is the declaration of the object's necessary behaviours, the second, implementation part contains the actual details of an object that we wish to use, that is to have the relevant characteristic behaviours. If the interface element defines the object in the abstract, the implementation part realizes it in the present concrete case. Since the algorithm relies on duck typing, the implementing object can be any type, as long as it accepts the functions in the interface command.

The theory of OOP also explains why interface elements are representatives of the complete object against outsider elements, which do not need to be cognizant of the object's details in order to be satisfied that such an object will be eventually supplied. According to OOP, interfaces are a kind of contract promising what the object can do once fully implemented. When the body of an object implements a certain interface, it fulfils the promise given by the declaration of behaviours in the interface. Thus, the interface guarantees a type of object by declaring that the object will support the set of identifying behaviours of that type, making the interface the object's representative (Meyer, 1988). Thus interface-implementation is a mechanism for satisfying the needs of the object's clients simply and immediately, while hiding the object's complexity for the time being.

We might now define this mechanism's two inputs, its output, and how it all works. The first input is a command for a behaviour which is specific and unique for a type of object; the second input is an actual element to which the command is to apply; the output is a typed object that has the defined features.

If we can show that a similar computational algorithm is applied in the FW-CW combinations in English, we have a mechanism for its generation of typed phrases and clauses. We know the English pattern has the right input and output. Function words are like interfaces: they are commands for applying various grammatical functions but have no semantic content. For instance, the article *the* confers definiteness on the content word it combines with but does not disclose anything about the content of the argument word. Content words are like implementations of the interface: they specify the details of the object to be defined, so that a common noun like *table* following the article *the* contains the semantics of the object thus defined. The outcome of the combination is a typed computational object which in our example is a determiner phrase (or noun phrase in traditional terms). Similarly, auxiliary verbs apply tense, aspect and mood features to verb-type predicate words. Subordinating conjunctions subordinate clauses. Relative pronouns define a predicate formed by lambda-abstraction, on the variable corresponding to the relative pronoun, of an open sentence (Montague, 1973). Interrogative wh-pronouns work similarly. Copulae add predicate value to any predicate expression following it, and so on. The specific content or meaning of their associates does not constrain the applicability of these general operations. As they do not have semantic content, we may say that grammatical words have their valency as their sole content. Their grammar is general, nonspecific; global rather than local; they are class-typical rather than instance-specific.

## 6. Testing the Fit of the Algorithm for English

To strengthen the conclusion that the English pattern indeed involves the processes of interfacing and implementation, we have tested it for the presence of features which are functional or critical for the proposed computational mechanism to work.

First and rather obviously, the algorithm dictates the temporal organization of the two moves. In OOP, An interface element occurs by definition as the first declaration of a new object, followed by the implementation. The order is functional: the implementation is of a specific interface, and it would be undefined if the interface were not already declared. The logical ordering predicts that the English pattern should also have a rigid word order, with the FW to the left of the CW, and also to the left of any of the CW's elaborators such as its adjuncts and complements. Checking the descriptive grammars, we find the prediction correct. The FW elements of English indeed occur in a rigid initial position, followed by the CWs (Hudson, 1990; Quirk et al, 1972). In addition, the word-order of the FW-CW unit of English is highly resistant to being altered by syntactic operations: the implementation-complement cannot be separated from the FW, cannot be moved by question movement, extraposition or topicalization (Cann, 2000). This is unusual in English as other syntactic combinations are mostly open to variations of ordering.

The literature on cross-linguistic patterns is clear about the word-order of a FW and the associated CW being inflexible not only in English but in languages in general. Bakker (1997) examined 137 European languages including English, and summarized that articles have almost completely frozen positions vis a vis the common noun; indefinite articles have zero flexibility; equally inflexible are the auxiliary verb and nonfinite verb, demonstrative and noun, adposition and noun, number and noun, relativizer and noun combinations. By contrast, word-combinations involving content words are more flexible, for instance the adjective-noun, verb-recipient object and verb-direct object combinations. In addition, even in otherwise completely free-word order languages, grammatical words tend to have a rigid word-order with their content word associates, and see for instance noun classifiers and nouns in Dyirbal (Dixon, 1972), Ancient Greek articles and postnominal attributes (Plank, 2003); or prepositions and nouns in Latin (Lehmann, 1991). These findings suggest the intriguing possibility that our proposal about the interface-implementation algorithm underlying the relevant FW-CW combinations is universally applicable.

The second functional feature of the interface-implementation mechanism in OOP is a strict differentiation between what is constant and what is variable in an object. The separation of interface and implementation follows the principle that all that is changing or variable is hidden inside the object, while all that is constant is visible at the front as a rigid interface. The interface is constant, essential and unchanging; all that changes is implementation. The principle, called *Protected Variation*, says: "Identify points of predicted variation and create a stable interface around them" (Cockburn, 1996). An almost identical principle is Meyer's (1988) *Open–Closed Principle* that says that objects should be open for extension but closed to modification that affects other objects. Extension in this context means changes to the internal details such as enlarging the range of possible behaviours, while modification is any change that would alter the nature of the stable public operations that define the object.

Protected Variation and the Open-Closed Principle are two expressions of the same root principle of OOP which is providing protection to the interface element from variations and changes while allowing them for the body of the object. According to Martin (2002), the principle is the answer to the question, "How can we create designs that are stable in the face of change?" This principle motivates many mechanisms and patterns in OOP that provide flexibility and protection from variations by minimizing the impact of change, such as data encapsulation, interfaces and polymorphism. By employing these mechanisms, the user of the object need to care only about the object's interface, not the implementations of it. This means the implementations can be changed without affecting the rest of the system. As Jacobson, Christerson, Jonsson and Overgaard (1992) say: the principle behind this strategy is "simply to get a more maintainable structure where changes will be local and thus manageable" (p. 185).

Checking descriptive grammars, we find that the FW and CW elements of English uphold this principle in many different ways.

First, FWs are historically or diachronically conserved. They belong to so-called closed classes to which no new items can normally be added, and that usually contains a relatively small number of items. They typically resist borrowing and other additions and changes such as neologisms, nominalization of verbs, derivational and other morphological word formation processes, and so on. By contrast, CWs belong to open classes that are open to change; as a consequence they have a large membership. As they may at any moment have a new member, the classes are technically infinite in size. For more detail, see Dixon (1977) and Quirk et al. (1972, pp. 44-47).

Second, FWs are typically unchanging in form: they keep their shape and do not undergo inflections and other morphological modifications. By contrast, CWs do undergo morphological modifications such as inflections, pluralization, comparatives and so on.

Third, FWs are typically constant in their syntactic function whereas CWs are notoriously heterosemous. Articles such as *an*, conjunctions such as *because*, demonstratives such as *this* and so forth usually have a unique syntactic use. A small number of FWs have more than one syntax, among them the demonstrative, subordinator and relativizer *that*, but these are the exceptions. Moreover, typically there is only a single FW word for a given syntactic function; as Emonds (1985) pointed out, closed class items tend to have no synonyms.

It follows that FWs but not CWs are extremely choosy in terms of their syntactic combinatory behavior. For instance, an English article is designed to add definiteness information to a common noun so that the combination can refer to some entity, definite or indefinite; it would be meaningless (as well as incorrect grammatically) to combine it with a verb or adverb that do not refer to anything. The outcome is that FWs are typically rigid in selecting their syntactic associates, choosing members of a very specific form-class, while CWs combine with several different types.

Fourth, FWs rigidly conserve their syntactic or selectional properties by not participating in such syntactic processes that change words' syntactic features such as passivization. By contrast, CWs do undergo such processes and hence their syntax is more varied.

Fifth, FWs rigidly keep their syntax and resist being "coerced" to change them by the type of CWs that attempt to combine with them, whereas CWs syntactic type can be altered when combining with a relevant FW (Pustejovsky, 1995). Moens and Steedman (1988, p. 14) introduced the term "coercion" to linguistics because, as they say, the phenomenon is loosely analogous to type-coercion in programming languages. Take note that in OOP interfaces "cast" more than type-coerce, namely, they choose one of existing alternative roles for an implementing object.

Sixth, FWs cannot take adjuncts, hence their meaning cannot undergo modification. CWs do take adjuncts, and their meaning is more variable.

Seventh, FWs' semantics cannot be coerced; they do not undergo so-called "semantic tailoring" by their dependents. CWs, by contrast, may vary their precise meaning as the result of being paired with various dependents (Allerton, 1982).

In summary, FWs are constant, unchanging, type-defining; CWs are specific, open to change, variable, flexible, and easy to modify. Their combination answers well to the separation of OOP's interface versus implementing elements.

The third functional feature of the interface-implementation mechanism is that being a 'duck typing' algorithm, it leaves open the details of the implementing object. It can be any element from any class of origin, as long as it is accepted as the implementation of the interface. It is said that interfaces are procedures to establish ad hoc "interface classes". That means that in OOP the role of inheritance is considerably diminished because the ad hoc "interface classes" override the objects' classes of origin. An interface allows objects belonging to unrelated parent classes to implement the same set of operations. This is one of the major uses of interfaces in OOP: Interfaces are used in programming situations when objects of several different parent classes may need to be allowed to respond to some 'method call'. Defining the shared functionality as an interface allows disparate objects to participate in the same computation as long as they correctly implement the interface. Once they do, the class of origin is overridden and becomes irrelevant. The implementing objects become equivalent.

The ability to create of ad hoc "interface classes" is a considerable strength of OO programming. Gamma et al. coined the mantra "Program to an interface, not an implementation". They add: "Don't declare variables to be instances of particular concrete classes. Instead, commit only to an interface defined by an abstract class." (Gamma et al., 1994, p. 18). Indeed, this mechanism frees computations from the constrains of objects' classes of origin, conferring a large degree of flexibility the programs would not have otherwise.

Testing English grammar, it is evident that its vocabulary poses the very computational situation for which 'duck typing' by interfaces is the solution in OOP. There are three basic issues that make form-class categories undefinable for English. First, English contains words with indeterminate syntactic behaviour like gerunds; a gerund like *inviting* is partly a verb (it gets objects) and partly a noun (it can be a subject) by its syntax (Chomsky, 1970; Hudson, 1990, p. 316-326). Theoreticians acknowledge their problematic nature for syntactic theory, for example Jackendoff (1977, p. 51) knows that they pose an unsolvable issue for X-bar theory.

The second problem is that it is in fact impossible to find groups of words that share all their grammatical behaviour (Crystal, 1967, p. 28; Levin, 1993, p. 18). Words do not have a single syntactic potential or subcategorization but several different ones. The combinatory profile of different words does not necessarily overlap. Crystal summarizes the problem by saying that if the criterion of overall identical syntactic behaviour were consistently applied, we would end up with a multitude of single member classes; Levin has a similar point. In addition, Sinclair (1999) claims that all common words of English have individual patterns of occurrence. Lyons (1968) describes the reasons why Chomsky (1965) gave up using word-classes based on distributional criteria in his grammar:

> It is possible to go a lot further with the distributional subclassification of words than would have been thought feasible, or even desirable, by traditional grammarians. But sooner or later, in his attempts to exclude the definitely unacceptable sentences by means of the distributional subclassification of their component words, the linguist will be faced with a situation in which he is establishing more and more rules, each covering very few sentences; and he will be setting up so many overlapping word-classes that all semblance of generality is lost. (p. 152-153).

The third problem concerns the extensive heterosemy found in the English lexicon. Heterosemy is the ambiguity of a word's formal potential. This is a feature of isolating and analytic languages, and, besides English, it has been extensively described for Chinese, Vietnamese, Thai and so forth (e.g., Enfield, 2006). A large part of the English vocabulary, basically most of the Anglo-Saxon core, consists of words that can serve as bona fide members of several major form-classes without any morphological-derivational marking. Words such as *kiss*, *talk* or *chase* can be used as verbs or nouns, many adjectives are also nouns, e.g., *green*, *sick*, or *poor*, or verbs, e.g. *wet*, *dry*, *dirty* and so forth. Called variously also conversion and zero derivation, this phenomenon has been extensively discussed in the linguistics literature, sometimes not merely as a problem impeding with the use of rigid form-class categories in grammar but even as the complete uprooting of the concept of words with a priori syntactic potentials. Thus Chomsky (2013, p. 43) speaks of "unspecified roots", Vogel (2000) of "ungrammaticalised lexemes" or "naked stems" and Borer (2005) of "classless listemes" and "exo-skeletal stem-allomorphs". This radical premise is obviously wrong as for no English word are all syntactic categories equally possible, rather, there is a probability distribution of a few possible ones, which demonstrates even the most heterosemous words are not truly type-neutral stems.

In present-day theories, the remedies proposed for the problem of heterosemy evoke some process by which the original form-class of lexemes is turned into a syntactically appropriate one. However, the inherent impossibility of defining form-class-like syntactic categories in the first place shows that the solution cannot be coercion or conversion into a member of an a priori defined form-class. The solution has to be that there are no a priori form-class values, rigid or ambiguous, which need to be either disambiguated or coerced into the present use. Instead, we claim, a syntactic mechanism exists that ensures that words fit the roles for which they are employed. That is the mechanism of 'duck typing' in which FWs serve as interfaces and the heterosemous CW items serve as implementations.

In our proposed model, the interface acts as a kind of sieve: all items which are able to undergo the grammatical combination of the interface element are accepted as the required data for the external head; it is unnecessary to either coerce or classify these words. What is relevant to this process is merely the potential range of combinations each word allows. In programming terms, the interface-implementation mechanism "casts" the ambiguous item into one of its potential uses. If the present use is possible, it will happen; otherwise, the combination will crash and will not take place. Namely, the very information that is considered ambiguous and problematic regarding the "underspecified" type-classification of these words is exactly sufficient for their functioning as implementation for an appropriate interface.

The one condition for the efficient functioning of the interface mechanism is, obviously, that the FWs of the language contain the precise behaviours that define the type of objects the clients (external heads) need. We know the English FW vocabulary is extremely well adapted to this role; determiners want the potential referential expressions to be able to receive a value for definiteness; auxiliary verbs want potential main predicates to receive tense, aspect, and mood, and so forth. These grammatical behaviours seem to fit well also the usual form-class categories; we could have written "noun" and "verb" easily in the previous sentence and get an almost-paraphrase of the Chomskian functional-category system. However it is easy to demonstrate that English FWs in fact work as interface-type definers of logical type, not of form-class. One of the most frequently used FW is the copula; however, the category it defines for its implementation complement is not in any form-class system. What is called in the literature the subject-complement role (XCOMP, as well as XADJ), are filled by every possible predicate term, be it a non-finite verb, an adjective, adverb or noun phrase. The definition is clearly of the logical type of a first-order predicate (c.f., Montague, 1973); finite verbs are excluded as the copula has already taken the place of the (single) inflected verb in the relevant sentence. Syntactic heads want logical types, not parts of speech. Using FWs as interface gives exactly that.

We might conclude that we have found a good fit between our programming model and the proposed mechanism in English grammar on three crucial features which are functional for such a mechanism in programming. English FW-CW combinations rigidly follow the necessary temporal organization of the purported interface and implementation moves; they obey the strict differentiation between what is constant (the FW) and what is variable (the CW) in an object defined by an interface; and English vocabulary presents the rampant heterosemy for which 'duck typing' by interfaces is the solution in programming. That is, the FW-CW combination not only contains inputs and output analogous to the Interface-implementation algorithm of OOP but it also presents this algorithm's major functional characteristics.

We have undertaken this study with the goal of discovering the way to a new theory of syntax that would possess psychological reality. The last question is, therefore, is there any evidence that the proposed interface-implementation mechanism is actually employed by people during the generation and interpretation of sentences? A thorough answer will have to wait for a dedicated program of research. The next section presents a review of available psycholinguistic evidence for the algorithm's psychological reality in processing, development and loss.

## 7. Psycholinguistic and Neurolinguistic Evidence On The Proposed Mechanism

### 7.1. Evidence on the facilitating effect of FWs on the processing of the phrase or clause

In section 5 we referred to the programming concept according to which the clients needing the computational object are satisfied by just the interface element as it guarantees that the right type of complete object is coming. In natural languages, the client of the object is a higher syntactic head demanding such an object as its complement, or allowing it as its adjunct. To be a contract for the whole object requires that the interface be an independent element of the algorithm, and that it is immediately interpreted and integrated with the preceding external head, rather than interpretation wait until the whole phrase or clause is completely received.

Processing data supports these predictions. Based variously on reading time, eye-tracking, and neuroscientific methods like the measurement of event-related potentials, it is consistently found that each word of the sentence is processed as soon as it is encountered, and it is immediately integrated with previously processed material into the hypothesized syntactic structure of the sentence (Crocker, Knoeferle, & Mayberry, 2010; Demberg & Keller, 2008;

Garrett, 1990; Marslen-Wilson, 1973; Pickering, 1999; and Tanenhaus, Spivey-Knowlton, Eberhard, and Sedivy, 1995).

Moreover, it was found that the FWs have facilitating effect on the cognitive load posed by the processing of the phrase or the clause. Some FWs are optional and the sentence is grammatical with or without them. Two main examples are subordinators (complementarizers) and some relative pronouns. In experimental studies it is found that phrases and clauses are processed faster and more accurately with these optional FWs than without them (Fodor & Garrett, 1967; Hakes, 1972; Hakes & Cairns, 1970). In addition, studies also reveal the process by which FWs may facilitate processing: words serve as the basis for the building of hypotheses about syntactic structures and semantic roles of the following input (Kuperberg & Jaeger, 2016; Kamide, 2008). The results of eye-tracking and neuropsychological studies thus support our model in which FWs serve as contracts predicting and guaranteeing the rest of the phrase.

### 7.2. Decisive role of FW-CW combinations in the development of syntactic competence

According to our model, FW-CW combinations are the central mechanism for generating syntactic constituents such as phrases and clauses. This predicts that the mastery of FW-CW combinations by young children have a decisive role in the development of syntactic competence. The results of several studies show that it is specifically the productive use of the FW vocabulary that predicts syntactic development (e.g., Le Normand, Moreno-Torres, Parisse, & Dellatolas, 2013; Szagun & Schramm, 2016). A recent special issue of *First Language* was devoted to this topic, presenting a series of studies exploring various aspects of this relationship (Ninio, 2019). As FWs do not appear as single-word utterances but only as part of a FW-CW combination, learning this vocabulary means learning the syntactic pattern it appears in.

Testing this learning model in experiments using miniature artificial languages showed that grammatical words indeed have a facilitating effect on adults' learning of the structure of the language (Gervain et al., 2013; Green, 1979; Morgan, Meier, & Newport, 1987; Valian & Coulson, 1988). A similar result was found for infants (Gervain, Nespor, Mazuka, Horie, & Mehler, 2008).

More specifically, Hallé, Durand, and de Boysson-Bardies (2008) demonstrated that articles help 11-month-old infants to recognize the following CWs. Moreover, Christophe, Millotte, Bernal, & Lidz (2008) showed that by two years of age infants can use FWs to infer the syntactic category of unknown CWs.

We may summarize that developmental patterns strongly suggests that the interface-implementation algorithm possesses psychological reality and plays a role in acquisition.

### 7.3. Loss of FWs correlates with loss of syntax

It is well documented that the loss of the FW vocabulary accompanies the loss of syntax in Broca's Aphasia following strokes and similar traumas (Goodglass & Kaplan, 1972), in Alzheimer's disease (Eyigoz, Mathur, Santamaria, Cecchi, & Naylor, 2020) and more. In these patients, the CW vocabulary does not show a similar damage. The high correlation between the functional vocabulary and the mastery of syntax that we saw in development, repeats in language loss, which we similarly interpret in terms of the constituents-building role of the FW vocabulary.

In summary, the literature on processing, development and loss is strongly supportive of the hypothesis that the FW-CW combination is a syntactic algorithm of the type interface-implementation, with psychological reality in processing, development and loss. Further experiments are needed to illuminate the details of the cognitive processing involved in this mechanism in children as well as adults.

## 8. Concluding Remarks

### *8.1. New insights on the syntax of English*

We have followed the call for a paradigm change in the cognitive sciences in general and formal linguistics in particular, and presented an attempt to describe the syntax of Modern English in the form of computational algorithms. The success of a new model is evaluated not only in terms of its fit to the empirical phenomena but also, and in particular, by its resulting in some unexpected new insights. The decision to use programming languages as heuristic models for natural languages yielded several such findings. We found a syntactic mechanism generating phrases and clauses with some as-yet undescribed computational roles: interfacing and implementation. Apparently the head-dependent relation is not the only one operative in structuring sentences; we should add abstraction-realization as well. With the help of this mechanism, we found out that English syntax uses a behaviour-based method for determining the type of phrases, namely 'duck typing'. This method not only solves in an elegant manner the problem of the widespread heterosemy in the vocabulary of English but also explains why the sheer concept of form-class is mostly unnecessary. That is good news as we have been repeatedly told that this concept is indefensible empirically. We also found that instead of FWs predicting a future item or a phrasal unit, we should better speak of their offering a contract guaranteeing them. Switching to a deterministic and algorithmic form of structuring sentences may make some further changes in theorizing. Interestingly, in one go we accounted for a long series of unusual features of grammatical words. We have known already that these words are historically conserved, have no synonyms, have a rigid position, content and shape, they cannot be moved, separated, changed, modified, coerced and so on. Their interfacing role explains all these and more. Grammars often call them 'operators' but the answer to the question what their operation actually consists of, needed to wait for the present model derived from programming languages.

Given the high usefulness of the interface-implementation mechanism, it is an important piece of the puzzle that English is not alone in utilizing such an algorithm. A quick search revealed that such a mechanism may be found in many other languages. A left-right, first-place interface mechanism is in particular characteristic of analytic languages (e.g., Thai, Vietnamese, Mandarin Chinese), but appears also in synthetic and even polysynthetic ones, see ditropic clitic determiners in Kwakwala, object clitics in Yagua and so on (Cysouw, 2005). An important variant are second position clitic clusters, which interface both the predicate and the referential elements of the clause (e.g. in Basque, Serbo-Croatian, or Warlpiri; see Anderson, 1993). A mechanism with a mirror image of its word order is the Japanese or Korean *bunsetsu* (Kurohashi & Nagao, 2003). It is out of the range of this article to discuss these and further relatives of the first-place interfacing mechanism; they are mentioned only to emphasize that a mechanistic syntactic theory may fit very well languages of other typological types besides English.

As the interface-implementation algorithm only covers about half of the syntactic relations of a given sentence, it should be made clear that English has, in addition, other programming-related  processing mechanisms such as coordination and gapping, or complex predicate formation, which are yet more sophisticated mechanisms for the introduction of computational objects. Similar to our interface-implementation algorithm, these other mechanisms are also manipulations of the logical type of input elements and their combinations. For instance, coordination and gapping in programming terms are computed by higher-order algorithms named 'add' and 'map2'. It is expected that it will not pose a considerable difficulty to express all of English syntax as algorithms derived from programming languages. That is the plan for a future project.

### 8.2. A positive contribution to client disciplines of linguistics

At the start of this paper, we described the need of the client disciplines of linguistics for a usable syntactic theory. Although the present study is merely the first step in the development of a theory describing likely cognitive mechanisms of syntax, it is possible to envision the potential of such a theory. Merely from considering the FW-CW combination, we have gained cognitive processes not yet considered in the context of syntax, that is, interfacing and implementation. A computing mechanism built on contracts and their fulfillment offers to cognitive disciplines a meaningful type of prediction built into the language that can replace the contingent or statistical concept of prediction often evoked in order to explain the interrelations between units of the sentence. In particular, such computational mechanisms may provide new explanations for well-documented phenomena of processing, development and loss of syntactic competence, as shown in the last section.

We also hope that the mechanism we have described and future discoveries of other algorithms will have a positive contribution to research aiming at understanding the neurobiology of syntactic processing. As summarized in a recent paper by Law and Pylkkänen (2021), despite decades of study, "the nature and even existence of purely structural processing in the brain remains elusive" (p. 2186). One of the major reasons for the failure is the straightforward refusal by theoretical linguistics to describe actual processes that human beings may carry out when generating and interpreting the syntactic structure of sentences, as we saw in Section 2. Despite some objections (e.g., Sutherland, 1966, and especially Pullum, 2009) the field of linguistics has been following suit ever since. It is time that this taboo is broken. Rules of grammar are derived from the actual behaviour of speakers and listeners, and reflect the computational processes people carry out. As long ago as 1977, Marr and Poggio already suggested that offering psychologically real cognitive algorithms to neuroscience is a required move towards the neuroscientific description of how cognitive processes are actually carried out. As a first step, we can offer our algorithm of contract and fulfillment as a possible concrete mechanism implementing the predictive nature of cognition and maybe also of the so-called *predictive brain* (Gilead, Trope, & Liberman, 2020), instead of statistical correlation. Assuming that cognition works through meaningful algorithms may be a more helpful way for finding neural correlates of it than the converse.

**Declaration of Conflicting Interest: none.**

**Acknowledgments**

I am grateful to Matan Ninio for lengthy and valuable discussions of Object Oriented Programming, and to Uri Hershberg for his input on the topic of modeling. Various aspects of the research were supported by the Center for Complexity Science [GR2007 043] and the Spencer Foundation [200900206].